\documentclass[runningheads]{llncs}

\usepackage{eccv}
\usepackage{float} 

\usepackage{eccvabbrv}

\usepackage{graphicx}
\usepackage{booktabs}
\usepackage{subcaption}
\usepackage{multirow}
\usepackage{colortbl}
\usepackage[accsupp]{axessibility}  

\usepackage{hyperref}

\usepackage{orcidlink}

\begin{document}

\title{Joint RGB-Spectral Decomposition Model Guided Image Enhancement in
Mobile Photography} 

\titlerunning{A Joint Decomposition Model Guided Image Enhancement}

\author{Kailai Zhou\inst{1} \and  
Lijing Cai\inst{1} \and
Yibo Wang\inst{1}  \and
Mengya Zhang\inst{1} \and 
Bihan Wen\inst{3} \and \\
 Qiu Shen\textsuperscript{$\dag$} \inst{1,2} \and
  Xun Cao\inst{1,2} }
\authorrunning{K Zhou et al.}

\institute{School of Electronic Science and Engineering, Nanjing University, Nanjing, China \and
Key Laboratory of Optoelectronic Devices and Systems with Extreme Performances of MOE, Nanjing University, Nanjing, China \and
Nanyang Technological University, Singapore \\
\email{\{calayzhou, cailijing, ybwang, 522023230161\}@smail.nju.edu.cn}, bihan.wen@ntu.edu.sg, \{caoxun, shenqiu\}@nju.edu.cn }



\maketitle
\footnotetext[1]{Corresponding author.}
\begin{abstract}
The integration of miniaturized spectrometers into mobile devices offers new avenues for image quality enhancement and facilitates novel downstream tasks. However, the broader application of spectral sensors in mobile photography is hindered by the inherent complexity of spectral images and the constraints of spectral imaging capabilities. To overcome these challenges, we propose a joint RGB-Spectral decomposition model guided enhancement framework, which consists of two steps: joint decomposition and prior-guided enhancement. Firstly, we leverage the complementarity between RGB and Low-resolution Multi-Spectral Images (Lr-MSI) to predict shading, reflectance, and material semantic priors. Subsequently, these priors are seamlessly integrated into the established HDRNet to promote dynamic range enhancement, color mapping, and grid expert learning, respectively. Additionally, we construct a high-quality Mobile-Spec dataset to support our research, and our experiments validate the effectiveness of Lr-MSI in the tone enhancement task. This work aims to establish a solid foundation for advancing  spectral vision in mobile photography.
The code is available at \url{https://github.com/CalayZhou/JDM-HDRNet}. 

\end{abstract}

\section{Introduction}
\label{sec:intro}







Recent advancements in miniaturized spectrometers \cite{yang2021miniaturization,mcgonigle2018smartphone} have facilitated  their integration into mobile devices, thereby expanding the scope of possible applications such as medical diagnosis \cite{he2020hyperspectral}, food safety evaluation \cite{fu2016food}, anti-spoofing face recognition \cite{rao2022anti} and color analysis \cite{finlayson2020designing}. Nevertheless, spectral sensors \cite{amsSpectralSensing} in mobile photography are  mainly employed for illuminant estimation in the Auto White Balance (AWB) operation at present. The spectral information holds potential to enhance the aesthetic appeal and perceptual quality of images. Exploring how to leverage additional spectral information to improve other image enhancement operations in the mobile Image Signal Processing (ISP) pipeline is an area requiring deeper investigation.

\begin{figure}[tb]
  \centering
  \includegraphics[width=0.9\linewidth]{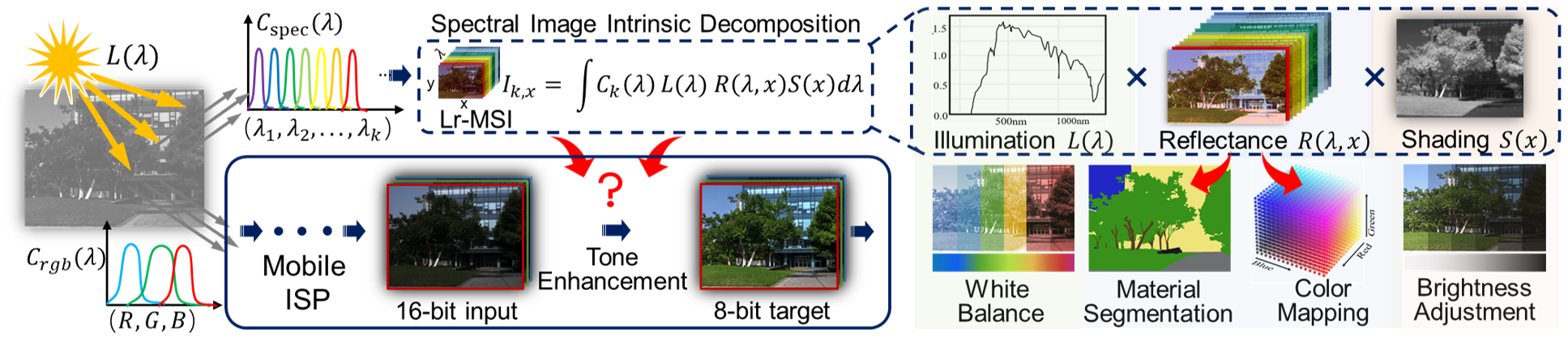}
  \caption{The Lr-MSI is decomposed into the illumination $L(\lambda)$,  reflectance $R(\lambda, x)$ and shading $S(x)$ terms. We then analyze their respective roles in the  mobile ISP pipeline.
  }
  \label{fig:introduction}
\end{figure}

We attempt to make a deeper analysis of the role of Lr-MSI in mobile photography by drawing insights from the spectral image intrinsic decomposition. As illustrated in \cref{fig:introduction}, the spectral image can be decomposed into its constituent terms \cite{chen2017intrinsic}: illumination curve $L(\lambda)$,  reflectance $R(\lambda, x)$, and shading $S(x)$. $L(\lambda)$ captures the spectral characteristics of the light source and has traditionally been applied for illuminant estimation \cite{erba2024computational}. $R(\lambda, x)$ embodies the albedo invariant color and texture of the material. The reflectance of spectral images which possesses fine-grained color channels shows promise for material segmentation and color mapping tasks. $S(x)$ represents interaction between the geometry of objects and illumination, which may contribute to the local brightness adjustment of high dynamic scenes.  We contend that the potential of the reflectance $R(\lambda, x)$ and shading $S(x)$ terms has not been thoroughly explored. 

In this paper, we explore the application of low-resolution multi-spectral images in the tone enhancement task within the mobile ISP pipeline, especially  for outdoor High Dynamic Range (HDR) scenes. The objective of tone enhancement is to adjust the brightness, contrast, and color characteristics of an image to enhance details in both the highlights and shadows while preserving its natural appearance. The primary challenge is the heterogeneous information fusion involving the integration of extra Lr-MSI with the original RGB workflow. This challenge manifests in two aspects: Firstly, variations in imaging mechanisms lead to the inherent complexity of spectral images, encompassing factors like scene geometry, inter-reflections and intricate artificial illumination \cite{chen2017intrinsic}, making it difficult to directly integrate spectral information into the mobile ISP workflow. Secondly, despite commercial spectral sensors offering over ten spectral channels \cite{amsSpectralSensing}, their spatial resolution is often confined due to the constrained spectral imaging capabilities on mobile devices. This limited spatial resolution remains a problem for the application of Lr-MSI in tone enhancement.


 To address the above challenges, we propose a joint RGB-Spectral decomposition model guided image enhancement framework which consists of two phases: joint decomposition and prior-guided enhancement. The effectiveness of this framework relies on three assumptions: (1) The near-infrared band in Lr-MSI can serve as an approximation of the shading term \cite{cheng2019non}. (2) Lr-MSI and RGB images exhibit complementary characteristics in both spatial and spectral resolutions. (3) The increased color channels in Lr-MSI contribute to material segmentation. Firstly,  to mitigate limited spectral imaging capabilities, the joint decomposition phase leverages the complementarity between Lr-MSI and RGB images to predict shading, reflectance, and material semantic priors. Secondly, to overcome the inherent complexity of spectral images, the prior-guided enhancement phase seamlessly integrates these priors into the established HDRNet \cite{gharbi2017deep} to provide clear mechanistic guidance. Specifically, the shading component is separated from original RGB images to enhance the adaptability in dealing with localized brightness variations. Subsequently, the $R(\lambda, x)$ of Lr-MSI is employed to guide the prediction of bilateral grids via the spectral perception map, particularly within the reflectance domain. Harnessing the reflectance domain alongside higher-dimensional spectral channels facilitates enhanced learning of color relationships for tone mapping. Lastly, we introduce the mixture of semantic grid experts to specifically adapt to the distinct color characteristics of individual material categories. The Joint DecoMposition framework guided HDRNet (JDM-HDRNet) exhibits superior performance compared to previous enhancement methods in terms of PSNR, SSIM \cite{wang2004image} and a color difference metric.



The contributions of this paper are as follows: (1) The pioneering Mobile-Spec dataset is meticulously constructed with high quality aligned RGB and hyperspectral images, which serves as a solid foundation for spectral applications. (2)  We propose a joint RGB-Spectral decomposition model guided enhancement framework to address challenges of inherent complexity of spectral images and limited spectral imaging capabilities. The Lr-MSI and RGB images are decomposed into shading, reflectance and material semantic priors, and these priors are employed for promoting dynamic range enhancement, color mapping, and semantic grid expert learning, respectively. (3) Experiments demonstrate the effectiveness of introducing extra Lr-MSI in the tone enhancement task, which provides valuable exploration for the industrial application of spectral sensors.




\section{Related Work}
\subsection{Potential of Mobile Spectral Sensors}

As the cost of miniaturized spectrometers \cite{yang2021miniaturization,mcgonigle2018smartphone} decreases and their availability becomes more widespread, the integration of these spectral sensors with mobile devices creates new opportunities. Studies have demonstrated the potential of hyperspectral imaging in medical applications \cite{he2020hyperspectral,ng2024hyper,he2019analysis,lindholm2022differentiating} such as hemodynamics monitoring \cite{he2020hyperspectral} and skin analysis \cite{ng2024hyper,he2019analysis}. Moreover, smartphones equipped with snapshot hyperspectral sensors can enhance anti-spoofing face recognition \cite{rao2022anti} by preventing  high-quality 3D mask attacks. Other applications include food safety evaluation \cite{fu2016food}, color analysis \cite{finlayson2020designing}, and water quality assessment \cite{e2022smartphone}. Despite these advancements, the use of spectral sensors in mobile photography remains limited. As RGB sensors lack inherent illumination adaptation abilities to maintain color constancy, in recent years, most high-end smartphones are equipped with spectral sensors which aim to improve the accuracy of illuminant estimation \cite{erba2024computational,thomas2015illuminant,glatt2024beyond} in an uncontrolled environment. The spectral sensor has evolved from its origins as an ambient light sensor to the refined capture of spectral signatures, primarily employed in auto white balance because its spatial resolution is often confined to a single pixel \cite{amsSpectralSensing}. Our objective is to validate the applicability of Lr-MSI to other enhancement tasks within the mobile ISP pipeline, particularly under assumptions where the constraints of spectral imaging are mitigated by dual-camera acquisition in the Mobile-Spec dataset.


\subsection{Enhancement in the Mobile ISP Pipeline}

A standard mobile ISP pipeline \cite{karaimer2016software} comprises both the Bayer processing routine and the subsequent photo-finishing routine. There are several enhancement procedures such as exposure correction \cite{yuan2012automatic}, white-balance \cite{gijsenij2011computational}, contrast enhancement \cite{cai2018learning}, color manipulation \cite{kim2012new}, local and global tone-mapping \cite{mantiuk2008display}. In this study, we aim to fully exploit the intrinsic decomposition components of spectral images by integrating Lr-MSI into the tone enhancement. This task involves adjusting the tonal range while preserving overall appearance and details, especially for HDR scenes. The commonly used method is the histogram equalization \cite{kim1997contrast} which globally improves the contrast of entire images. However, global operations often fall short when dealing with scenes of high dynamic range. In the deep learning era, color transform-based methods are often employed for real-time processing of high-resolution images in a collaborative global-local manner. According to the color transform functions, these methods can be classified into affine transformation matrices \cite{wang2019underexposed,gharbi2017deep}, multilayer perceptrons (MLPs) \cite{he2020conditional}, 3D LUTs \cite{zhang2022clut,yang2022seplut,zeng2020learning}, etc. The pioneering work is the HDRNet \cite{gharbi2017deep} which learns the bilateral grid of coefficients at low resolution and performs transformations from input to output at full resolution. 3D LUT \cite{zeng2020learning} replaces the affine transformation matrices with image-adaptive 3-dimensional lookup tables, enabling rapid and robust photo enhancement. We choose the HDRNet as the baseline due to its simple and flexible network design. The key challenge lies in the heterogeneous information fusion involving the integration of extra Lr-MSI with the HDRNet.

\subsection{Spectral Image Intrinsic Decomposition}



To fuse the extra Lr-MSI into the original RGB workflow in HDRNet, we consider the role of spectral information in mobile photography from the perspective of Spectral Intrinsic Image Decomposition (SIID) \cite{chen2017intrinsic}. The SIID problem aims to separate a spectral image into more elaborate intrinsic properties, such as the spectrum of the illumination, reflectance and shading components. This decomposition is typically performed in the spectral domain, which allows much more flexibility in the algorithm design and leads to a more accurate analysis. Zheng et al. \cite{zheng2015illumination} model illumination and reflectance spectra separation of a hyperspectral image into a low-rank matrix factorization problem. Chen et al. \cite{chen2017intrinsic} resolve a spectral image into independent intrinsic components: illumination, shading, and reflectance, and develop an effective algorithm which first reduces the spatial and spectral resolution with the super-pixel down-sampling, then the SIID problem is optimized under local constant and global sparse constraint. Huang et al. \cite{huang2018multispectral} extend the Retinex model to the multispectral domain based on the subspace constraint, which assumes the reflectance and shading vectors both live in a low dimensional subspace along the spectral domain. SIID is an ill-posed and under-constrained problem and previous algorithms are developed under the ideal lab environment, underscoring the challenges of tackling complex outdoor scenes.  Fortunately, Cheng et al. \cite{cheng2019non} find that near-infrared images can function as reliable approximations of shading priors, and Zhang et al. \cite{zhang2022hsi} further perform intrinsic decomposition for outdoor scenes using aligned RGB and hyperspectral images. Based on this assumption, our joint  decomposition model can be trained in an end-to-end manner, which will be discussed in Section \ref{sec41}. 

\begin{figure}[tb]
  \centering
  \includegraphics[width=0.9\linewidth]{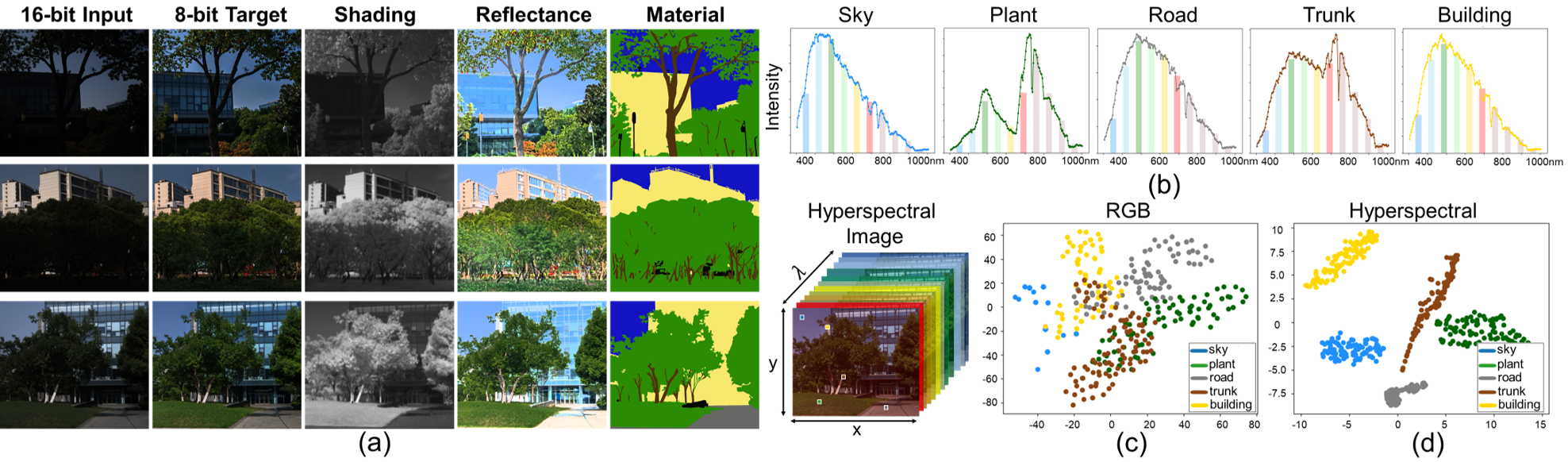}
  \caption{(a) The Mobile-Spec dataset comprises RGB images (16-bit input and 8-bit target), hyperspectral images and their corresponding shading, reflectance and material segmentation images. Near-infrared images serve as the guide map to approximate the shading term.  (b) Spectral responses of different material categories. (c-d) Visualizations of RGB and hyperspectral data for different categories with t-SNE \cite{van2008visualizing}.
  }
  \label{fig:dataset}
\end{figure}

\section{The Mobile-Spec Dataset}

To the best of our knowledge, no existing image enhancement dataset includes both aligned hyperspectral and RGB images. To bridge this data gap, we have constructed the high-quality Mobile-Spec dataset, consisting of 200 image groups. As shown in \cref{fig:dataset}(a), each group represents an independent scene and includes a 16-bit input image, an 8-bit target image, a hyperspectral image, and the corresponding shading, reflectance, and material segmentation images. 

\textbf{Dual Camera System.}  To address the limited spectral imaging capabilities of sensors on smartphones, the Mobile-Spec is captured using a dual camera system: a high-end commercial smartphone and a Gaiasky-mini2 scanning hyperspectral camera \cite{dualixGaiaskyminiHyperspectral}. The Gaiasky-mini2 covers a spectral range from 400 nm to 1000 nm with 176 spectral channels. The 16-bit input images are  fused from multiple exposure frames, and the 8-bit target images are generated by a commercial privacy model that integrates the aesthetics of multiple experts.
 

\textbf{Image Matching.} The hyperspectral image size is $1057 \times 960 \times 176$, while the 16-bit input and 8-bit target RGB images are $4096 \times 3072 \times 3$. To align  RGB images with a wide angle of view onto hyperspectral images, we employ an automatic alignment algorithm that computes SIFT descriptors \cite{lowe2004distinctive} and estimates homography transformation matrices. Consequently, both RGB and hyperspectral images achieve a consistent spatial resolution of $1057 \times 960$.

\textbf{Influence of Light Sources.} The Mobile-Spec dataset is captured exclusively outdoors to avoid the impact of intricate indoor lighting. Solar radiation spans a broad range of wavelengths, ranging from ultraviolet to visible light and extending into the infrared region. It is reasonable to assume that near-infrared images can approximate shading priors in outdoor settings \cite{cheng2019non}.

\textbf{Dataset Collection and Filtering.} Our dataset spans multiple seasons and  includes five categories commonly observed in outdoor environments: sky, building, plant, trunk, and road. Their respective spectral curves are illustrated in \cref{fig:dataset}(b). Mobile-Spec undergoes rigorous filtering to ensure high quality. Firstly, samples are selected based on dynamic range, determined by an automated algorithm analyzing the difference between maximum and minimum pixel intensities. Secondly, samples exhibiting significant alignment errors are filtered out. Thirdly, samples with unsatisfactory visual quality are excluded, considering factors such as chromatic aberration, sharpness, noise, and artifacts. Lastly, to augment dataset diversity, scenes with high similarity are eliminated.

\textbf{Shading, Reflectance and Material Segmentation.} To obtain targets of $S$, $R$ and $M$ for the joint decomposition model, shading images are approximated by averaging near-infrared bands from 850 nm to 1000 nm in hyperspectral images. Reflectance images can be derived based on the Retinex theory \cite{barrow1978recovering,land1971lightness}. Ground truths of material segmentation are meticulously labeled by human annotators. More details about the Mobile-Spec  are provided in the appendix.


\section{Proposed Method}

\subsection{Joint RGB-Spectral Decomposition Model}
\label{sec41}
To integrate low-resolution spectral images into the RGB tone enhancement workflow, we first  analyze the differences between hyperspectral and RGB camera imaging mechanisms. The imaging model for spectral cameras can be represented as follows based on \cite{jiang2013space,chen2017intrinsic}:

\begin{equation}
I_{k, x}=\int_{400 \mathrm{~nm}}^{1000 \mathrm{~nm}} C_k(\lambda) L(\lambda) S(x) R(\lambda, x) d \lambda, \mathrm{k}= \text{1, 2, 3 \ldots}  
\end{equation}

$I_{k, x}$ represents the pixel intensities at position $x$ of the k-th band, $C_k\left(\lambda\right)$ denotes the camera sensitivity function, $L(\lambda)$ indicates the illumination curve of the light source, $S(x)$ refers to shading, and $R(\lambda, x)$ is the reflectance of scenes. $L(\lambda)$ is usually utilized for illuminant estimation in the auto white balance procedure, while the role of the shading $S(x)$ and reflectance $R(\lambda, x)$ components has been rarely explored. The core difference between the RGB and hyperspectral cameras lies in $C_k\left(\lambda\right)$, where hyperspectral images have fine-grained color channels due to a higher spectral sampling rate. To simulate the limited spectral imaging capability on smartphones, we downsample the  hyperspectral image ($1057 \times 960 \times 176$)  to the Lr-MSI ($16 \times 16 \times 10$). This downsampling is conducted to match the typical 10-channel configuration of commercial smartphone spectral sensors. \cite{amsSpectralSensing}. We maintain flexibility in setting the spatial resolution to a default of $16 \times 16$, which will be discussed in Section \ref{exp52}. We posit that the effect of Lr-MSI in the joint decomposition model can be observed in two aspects.
   
\textbf{Approximate estimation of shading priors with the near-infrared band.} To obtain the shading term, traditional optimization-based methods design various hand-crafted priors to constrain  the solution \cite{chen2017intrinsic,huang2018multispectral}, which are often inapplicable in complex outdoor scenes. Recently, near-infrared images have offered a simple yet powerful approach for shading smoothness \cite{cheng2019non}, as the spectral curves of different colors gradually flatten out and texture variation is considerably reduced in the near-infrared band (refer to the appendix). Given that the Mobile-Spec dataset is exclusively captured outdoors to avoid the impact of complex indoor light sources, and considering the wide spectrum of solar radiation covering  the infrared region, the shading term can be predicted through the joint decomposition of Lr-MSI and RGB images, as they share the same $S(x)$.

\textbf{Enhancing material segmentation with the visible band.} We employ t-SNE \cite{van2008visualizing} to analyze the complex spectral patterns and relationships between different spectral bands. In \cref{fig:dataset}(c), clusters representing different categories exhibit denser aggregation in the t-SNE visualization of hyperspectral images compared to that of RGB images. This observation suggests that the increased granularity of spectral channels provides enhanced discriminative capacity among various materials, and Lr-MSI may potentially contribute to the material segmentation task.

\begin{figure}[t]
  \centering
  \includegraphics[width=1.0\linewidth]{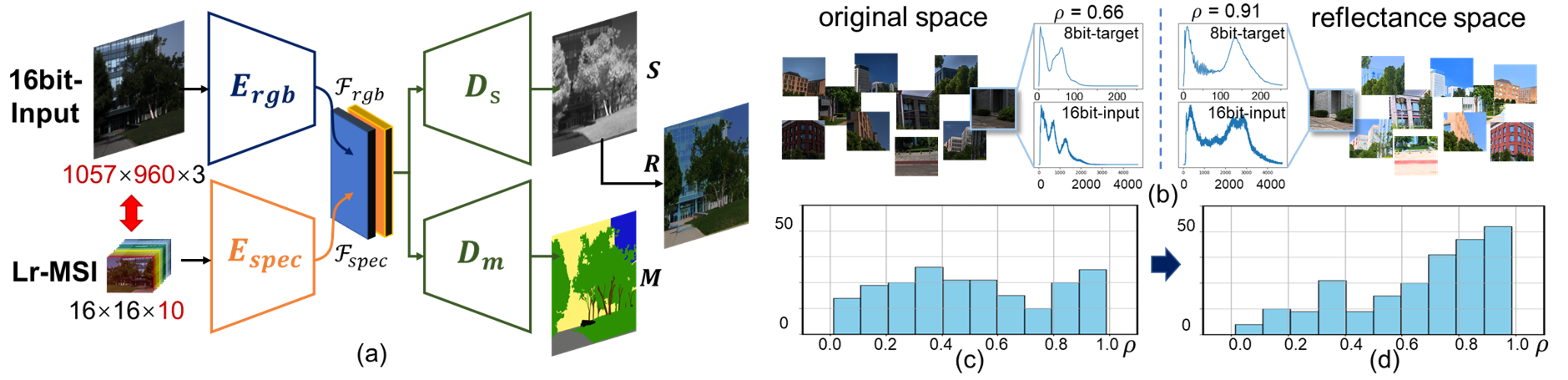}
  \caption{(a) The joint RGB-Spectral decomposition model leverages
the complementarity between RGB images and Lr-MSIs to predict
shading (S), reflectance (R) and material category (M) priors.  (b) The pixel intensity histogram. (c-d) The distribution of Pearson correlation coefficient $\rho$ in the original RGB space and reflectance space.
  }
  \label{fig:jdecoup}
\end{figure}

In \cref{fig:jdecoup}(a), we employ a joint RGB-Spectral decomposition model to leverage both near-infrared and visible bands in Lr-MSI for shading prediction and material segmentation. Lr-MSI provides high spectral resolution but limited spatial resolution, while the RGB image offers rich spatial information but lower spectral resolution. We transform the prediction of shading images from the regression problem to the classification problem by dividing shading images into eight brightness levels. Consequently, shading prediction and material segmentation can be decomposed in a collaborative manner. The joint RGB-Spectral decomposition model incorporates two independent encoders and decoders, adapted from the basic segmentation model FCN \cite{long2015fully}. Specifically, Lr-MSI is resized to match the spatial resolution of the RGB image.  The RGB encoder $E_{rgb}$ and spectral encoder $E_{spec}$ are trained to project the 16-bit input image $I_{rgb}$ and Lr-MSI $I_{msi}$ to the same latent space for representation alignment: $\mathcal{F}_{rgb}=E_{rgb}(I_{rgb})$ and $\mathcal{F}_{spec}=E_{spec}(I_{msi})$. $\mathcal{F}_{rgb}$ and $\mathcal{F}_{spec}$ are fused together with the concatenation to share the common and complementary feature representations. Material segmentation $M$  and shading $S$ are independently predicted with the decoder $D_{m}$ and $D_{s}$ based on the fused feature representation. The joint RGB-Spectral decomposition model can be formulated as:

\begin{equation}
M, S = D_{m,s}(concat(\mathcal{F}_{rgb},\mathcal{F}_{spec}))
\end{equation}

With the shading component $S$, the reflectance component $R_{rgb}$, $R_{msi}$ can subsequently be obtained according to the Retinex theory \cite{barrow1978recovering,land1971lightness}.

\subsection{JDM-HDRNet}
 Based on the established HDRNet \cite{gharbi2017deep}, the JDM-HDRNet comprehensively  exploits the $S$, $R$ and $M$ priors to provide explicit guidance for tone enhancement.

\textbf {$S$: Localized Brightness Adaptation.}  To adjust both highlight and shadow areas during tone enhancement, the HDRNet's bilateral grid depth is set to 8, accommodating eight different luminance levels. However, this design may not effectively handle various localized brightness variations in high dynamic range scenes. We propose that separating the shading component would yield benefits in such contexts. As shown in \cref{fig:jdecoup}(b), the pixel histogram vectors between the 16-bit input and 8-bit target exhibits a Pearson correlation coefficient $\rho$ of 0.66 in the original RGB space, which increases to 0.91 in the reflectance space. A larger $\rho$ indicates greater similarity in pixel histogram characteristics. Analyzing the statistics of the $\rho$ distribution in the Mobile-Spec dataset, we observe that the reflectance space (\cref{fig:jdecoup}(d)) has a greater proportion of samples with high $\rho$ than the RGB space (\cref{fig:jdecoup}(c)). This finding indicates that separating the shading component from the RGB space to the reflectance space may reduce the difficulty of color mapping learning. In our implementation, the shading component $S$ is converted to the brightness representation $\hat{S}$ with a localized brightness adaptation module, comprising two layers of convolution and deconvolution. This lightweight module helps  mitigate drastic pixel intensity changes in localized high dynamic range areas, thereby enhancing the adaptability. Subsequently, the reflectance images of 16-bit input $R_{rgb}=I_{rgb}/\hat{S}$  and Lr-MSI $R_{msi}=I_{msi}/\hat{S}$ are fed into the coefficient prediction part of the bilateral grid.

\begin{figure}[t]
  \centering
  \includegraphics[width=1.0\linewidth]{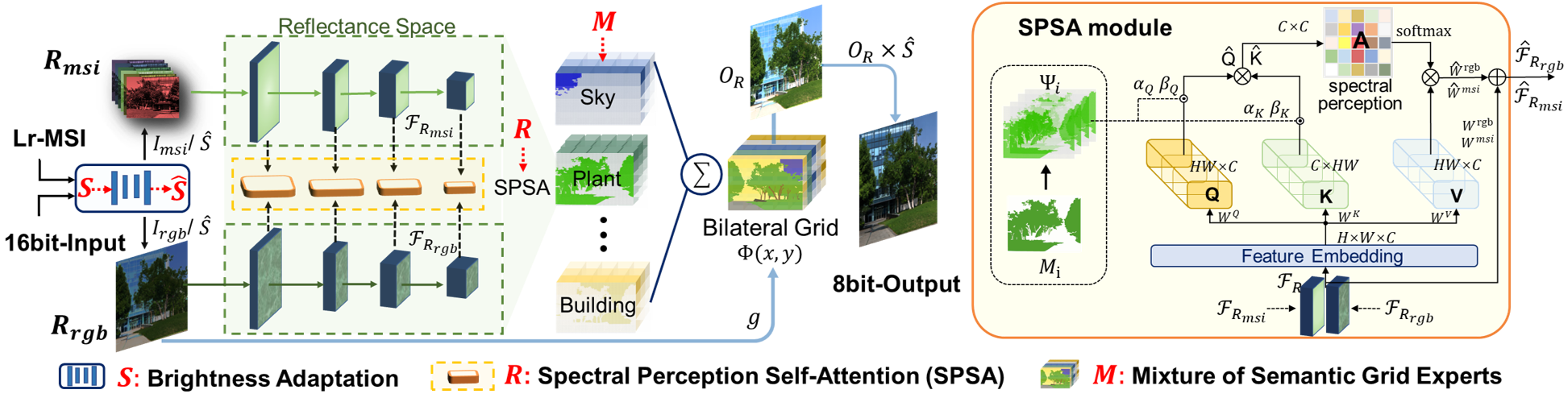}
  \caption{The overall architecture of the proposed JDM-HDRNet. The $S$ prior is employed  to enhance localized dynamic ranges, while $R$ and $M$ priors contribute to the color mapping and grid expert learning, respectively.
  }
  \label{fig:method}
\end{figure}

\textbf {$R$: Spectral Perception Self-Attention.}  After transforming from RGB space to reflectance space, we aim to leverage the reflectance of Lr-MSI $R_{msi}$ which has fine-grained spectral channels to enhance the color mapping learning of bilateral grid. It should be noted that $R_{msi}$ has limited spatial resolution. We argue that downsampling in the spatial dimension will not cause significant degradation since bilateral grid coefficient prediction is carried out at a much lower spatial resolution. Hence, we design the Spectral Perception Self-Attention (SPSA) module which utilizes the $R_{msi}$ to guide $R_{rgb}$ for  better bilateral coefficient prediction. $R_{msi}$ is resized to match the spatial resolution of $R_{rgb}$,  which is $256 \times 256$ by default in HDRNet. The $R_{msi}$ and $R_{rgb}$ are processed with a stack of strided convolutional layers to extract the low-level features and generate hierarchical feature maps. These feature maps $\mathcal{F}_{R_{msi}}$, $\mathcal{F}_{R_{rgb}}$ are concatenated along the channel dimension to form the cross-spectral feature embedding  $\mathcal{F}_R$. Then $1 \times 1$ convolutions $W_{1}$ and $3 \times 3$ depth-wise convolutions $W_{3}$ are applied to the $\mathcal{F_R}$ to generate query $Q$, key $K$ and value $V$: 

\begin{equation}
 {Q=W_{1}^Q W_{3}^Q\mathcal{F_R}},{K=W_{1}^K W_{3}^K\mathcal{F_R}},{V=W_{1}^V W_{3}^V\mathcal{F_R}}
\end{equation}

The query $Q$ and key $K$ are reshaped into $\hat{Q}\in\mathbb{R}^{HW\times C}$ and $\hat{K}\in\mathbb{R}^{C\times H W}$ so that their dot-product interaction generates a spectral perception map $A\in\mathbb{R}^{C\times C}$, which models the mutual information across the different spectral channels. The importance of different channels of the reshaped value $\hat{V}\in\mathbb{R}^{H W\times C}$ is reweighted by $A$, and the SPSA module can be formulated as follows:

\begin{align}
A &= softmax(\sigma\hat{K}\cdot \hat{Q}) \\
\hat{\mathcal{F}}_{R_{msi}} &= \hat{W}_{3}^{msi} \hat{V} \cdot A + W_{3}^{msi}\mathcal{F}_{R_{msi}} \\
\hat{\mathcal{F}}_{R_{rgb}} &= \hat{W}_{3}^{rgb} \hat{V} \cdot A + W_{3}^{rgb}\mathcal{F}_{R_{rgb}} 
\end{align}

The $\sigma$ is a learnable scaling parameter to control the magnitude of the dot product. The SPSA module serves as the residual learning which fuse the cross-spectral feature $\hat{V}\cdot A$ with the original feature $\mathcal{F}_{R_{msi}}$, $\mathcal{F}_{R_{rgb}}$ by the adaptive weighting of $3\times 3$ convolution $\hat{W}_{3}^{msi}$, $\hat{W}_{3}^{rgb}$. The fused outputs $\hat{\mathcal{F}}_{R_{msi}}$, $\hat{\mathcal{F}}_{R_{rgb}}$ of the SPSA module are fed into next layer in a progressive way. 







\textbf {$M$: Mixture of Semantic Grid Experts.}  Common outdoor scenes can be roughly categorized into several primary types (e.g., sky, building, plant). As illustrated in \cref{fig:dataset}(b), the plant spectral curves typically feature a reflection peak at around 550 nm, whereas the spectral curve of sky primarily resides within the blue wavelength range. Given these distinct characteristics, a multi-expert design can be introduced to effectively adapt to each material category's color preference. We believe that incorporating the material semantic prior $M$ can offer robust constraints and yield visually pleasing results for tone enhancement. Consequently, we introduce a mixture of semantic grid experts, wherein each category is individually associated with a unique bilateral grid coefficient. For a specific semantic grid expert, the binarized material segmentation of the $i$-th category $M_i$ is transformed into the probability map $\Psi_i$ with a mapping function comprising two convolutional and deconvolutional layers. Subsequently, a pair of affine transformation parameters $(\alpha,\beta) \in\mathbb{R}^{HW \times C}$ is learned based on the probability map  $\Psi_i$. The $(\alpha,\beta)$ is then introduced to the $Q$ and $K$ branch to modulate features conditioned on the probability map as follows:

\begin{equation}
\hat{Q} = (1+\alpha_Q) \cdot Q + \beta_Q, \hat{K} = (1+\alpha_K) \cdot K + \beta_K
\end{equation}

In each semantic grid expert $\Phi_i(x, y)$, the material category $M_i$ is introduced in the same way. The bilateral grids of different experts are dynamically fused based on the category-dependent weight coefficient denoted as $w_i$:

\begin{equation}
\Phi(x, y) = \sum_{i=1}^N w_i \Phi_i(x, y)
\end{equation}
 Here, $N$ is the number of material categories, and $\Phi(x, y)$ denotes the final multi-expert bilateral grid. The full-resolution output of reflectance image $O_R$ is obtained through interpolated affine transformations, which is operated based on the bilateral grid $\Phi(x, y)$ and the guidance map $g$. The $g$ is learned from the $R_{rgb}$ using the original guidance map auxiliary network. With the multiplication of $O_R$ and $\hat{S}$, we get the final 8-bit output of JDM-HDRNet.

\section{Experiments}
\subsection{Implementation Details}
In the Mobile-Spec dataset, $80\%$ of the samples are allocated for the training set, while the remaining $20\%$ are designated for the test set.  The JDM-HDRNet is trained with a batch size of 4 and a learning rate of 0.0001. Specifically, the joint decomposition model is first trained with cross-entropy loss for both shading and segmentation predictions, then the JDM-HDRNet is trained with MSE loss in a supervised manner. During the training, input images are cropped to the size of $512\times512$, and these images undergo further downsampling to $256\times256$ in the low-resolution stream.

Three metrics are employed to validate the effectiveness of our method, including the PSNR, SSIM \cite{wang2004image}, and $\triangle E^\ast$. $\triangle E^\ast$ measures the perceptual difference between two colors in the CIELAB space, a smaller $\triangle E^\ast$ value indicates better color accuracy or image quality. For the evaluation of the material segmentation, we adopt the standard metric as mean Intersection-over-Union (mIoU).


\begin{table}[t]
\setlength{\abovecaptionskip}{0.0em}
\setlength{\belowcaptionskip}{0.0em}
\caption{Performance evaluation of HDRNet with shading $S^{\ast}$ and reflectance prior $R^{\ast}$. Ablation studies are conducted on the spectral channel and spatial resolution.}

\centering
\begin{subtable}{0.9\textwidth}
\centering
\caption{Performance evaluation of HDRNet and HDRNet with shading prior (HDRNet+$S^{\ast}$). }
\resizebox{!}{0.41cm}{
\setlength{\tabcolsep}{1.5mm}{
\begin{tabular}{cccc|cccc}
\cline{1-4} \cline{5-8}
       & PSNR↑   & SSIM↑  & $\triangle E^\ast$↓    &     & PSNR↑  & SSIM↑  & $\triangle E^\ast$↓    \\ \cline{1-4} \cline{5-8} 
HDRNet & 27.75 & 0.939 & 5.12 &  HDRNet+$S^{\ast}$ & 28.68 & 0.957 & 4.29 \\ \cline{1-4} \cline{5-8}
\end{tabular}
}
}
\end{subtable}

\begin{subtable}{0.45\textwidth}
\centering
\caption{Ablation study of reflectance prior $R^{\ast}$ on channels with different spectral ranges.}
\centering
\resizebox{!}{1.5cm}{
\setlength{\tabcolsep}{1.2mm}{
\begin{tabular}{c|ccc}
\hline
 & PSNR↑  & SSIM↑  & $\triangle E^\ast$↓    \\ \hline 
baseline                                                   & 28.68 & 0.957 & 4.29 \\ \hline
400-520 nm                                                          & 28.74 & 0.957 & 4.00  \\
520-640 nm                                                          & 29.09 & 0.964 & 3.82 \\
640-760 nm                                                        & 29.19 & 0.962 & 3.92  \\
400-760 nm                                                        & 29.24 & 0.967 & 3.72 \\
760-1000 nm                                                        & 29.07 & 0.965 & 3.88 \\
\rowcolor[HTML]{EFEFEF} 
400-1000 nm                                                   & 29.68 &0.968 & 3.55 \\ \hline
\end{tabular}
}
}
\end{subtable}
\begin{subtable}{0.45\textwidth}
\centering
\caption{Ablation study of reflectance prior $R^{\ast}$ on the spatial resolution, which is ranging from $1\times 1$ to $256\times 256$.}
\centering
\resizebox{!}{1.35cm}{
\setlength{\tabcolsep}{1.5mm}{
\begin{tabular}{c|ccc}
\hline
 & PSNR↑  & SSIM↑  & $\triangle E^\ast$↓    \\ \hline 
baseline                                                   & 28.68 & 0.957 & 4.29 \\ \hline
$1\times 1$                                                        & 29.05 & 0.963 & 4.06 \\
$4\times 4$                                                         & 29.21 & 0.963 & 3.80 \\
\rowcolor[HTML]{EFEFEF} 
$16\times 16$                                                         &29.68 & 0.968 & 3.55  \\ 
$64\times 64$                                                        & 29.81 & 0.969 & 3.47 \\
$256\times 256$                                                  & 29.78 & 0.967 & 3.49 \\ \hline
\end{tabular}
}
}
\end{subtable}

\label{tab:abla_s_r}
\end{table}


\subsection{Effectiveness of $S^{\ast}$, $R^{\ast}$, $M^{\ast}$ Priors}
\label{exp52}

To avoid the impact of inaccurate decomposition, we begin by systematically integrating the ideal values of these priors (denoted as $S^\ast$, $R^\ast, M^\ast$) individually into the HDRNet baseline for validation.

\textbf{Shading prior $S^{\ast}$.} \cref{tab:abla_s_r}(a) illustrates that the HDRNet baseline attains a PSNR of 27.75 dB, while the separation of the shading component ($S^{\ast}$) elevates the PSNR from 27.75 dB to 28.68 dB. Notably, the architecture of HDRNet remains unchanged in this sub-experiment. This finding demonstrates that transforming from RGB into reflectance space of input images with $S^{\ast}$ is a simple yet effective design for the tone enhancement task, which exhibits enhanced adaptability in diverse  high  dynamic range scenes.  


 
\begin{table}[t]
\setlength{\abovecaptionskip}{0.0em}
\setlength{\belowcaptionskip}{0.0em}
\centering
\caption{Performance evaluation of HDRNet with material semantic prior $M^{\ast}$. }
\begin{subtable}[t]{0.55\textwidth}
\caption{The effectiveness of incorporating extra Lr-MSI in the joint RGB-Spectral decomposition model for the material segmentation task. }
\resizebox{!}{0.62cm}{
\setlength{\tabcolsep}{0.50mm}{
\begin{tabular}{c|ccccc|c}
\hline
          & building  & plant  & sky   & trunk & road  &  mIoU \\ \hline
RGB       & 81.99   & 87.63 & 94.84 & 10.96 & 83.87 & 71.86  \\
+LrMSI & 87.53 & 91.01 & 95.87 & 31.14 & 89.11 &   78.93   \\ \hline
\end{tabular}
}
}
\end{subtable}
\begin{subtable}[t]{0.4\textwidth}
\centering
\caption{Ablation study of $M^{\ast}$ on the number of material categories. }
\resizebox{!}{0.770cm}{
\setlength{\tabcolsep}{1.0mm}{
\begin{tabular}{c|cccc}
\hline
     & 1    & 2    & 4    & 6    \\ \hline
PSNR↑ & 27.75& 27.98 & 28.47 & 29.11   \\
SSIM↑ & 0.939 & 0.944 & 0.954 & 0.964  \\
$\triangle E^\ast$↓    & 5.12  & 5.35  & 4.69 & 4.28 \\ \hline
\end{tabular}

}
}
\end{subtable}
\label{tab:abla_m}
\end{table}

     
\textbf{Reflectance prior $R^{\ast}$.}  Given the practical limitations of spectral imaging capability on mobile devices, we conduct ablation studies to examine  the spectral and spatial resolution configurations of Lr-MSI. We first maintain the spatial resolution at the default setting of $16 \times 16$ and extract channels of different spectral ranges. The Lr-MSI has ten spectral channels, ranging from 400 nm to 1000 nm, with each channel featuring a 60 nm bandwidth. Notably, 400-760 nm encompasses six visible wavelength channels, while 760-1000 nm includes four near-infrared channels. As observed in \cref{tab:abla_s_r}(b), the closer to the wavelength of red channel, the higher PSNR it will achieve. The possible explanation lies in the composition of our Mobile-Spec dataset, where a larger proportion of the area is covered by the sky and plant, while the presence of red objects is comparatively limited. Consequently, the HDRNet baseline tends to prioritize the blue and green wavelengths over the red wavelength during the training. Furthermore, the combination of all ten channels (400-1000 nm) achieves the best results of 29.68 dB. In our view, fine-grained spectral information enhances bilateral grids with heightened color perception capabilities and compensates for the imbalanced learning of different color channels. Then we maintain the number of channels at ten, and the spatial resolution is resized from $1 \times 1$ to $256 \times 256$. As depicted in \cref{tab:abla_s_r}(c), there is a trend of improved performance in evaluation metrics with higher spatial resolution. Considering the inherent constraint of spatial resolution in Lr-MSI on mobile devices, we contend that the optimal parameter configuration for Lr-MSI may align to the spatial resolution of the bilateral grid ($16 \times 16$), which is the 8x downsampling corresponding to the spatial resolution of low-resolution stream input ($256 \times 256$).  Our experimental results suggest that the $16 \times 16$ configuration yields satisfactory results, beyond which increasing the spatial resolution leads to diminishing marginal utility.

\textbf{Material semantic prior $M^\ast$.} \cref{tab:abla_m}(a) demonstrates that incorporating the Lr-MSI branch into the joint RGB-Spectral decomposition model enhances the mIoU of material segmentation from 71.86\% to 78.93\%. Despite Lr-MSI's notably lower spatial resolution compared to the 16-bit, it exhibits superior material identification capability owing to its fine-grained spectral channels. Subsequently, we conduct ablation experiments to evaluate the number of material categories. The category number of the original HDRNet baseline is set to one. In the two-category set, the sky, plant, and trunk are grouped into one category, while building, road, and others form another. For the four-category set, the plant and trunk, as well as building and road, are each combined into single categories, whereas the sky and others are treated as separate, independent categories. By refining the division of material categories, the mixture of semantic grid experts becomes more specialized in enhancing tones for specific materials. \cref{tab:abla_m}(b) demonstrates improved quantitative results across three evaluation metrics with the inclusion of six semantic grid experts.

\subsection{Ablation Study of JDM-HDRNet}
\cref{tab:jdm_hdrnet}(a) presents the ablation study incorporating ideal values $S^\ast$, $R^\ast$, $M^\ast$ into HDRNet (denoted as JDM-HDRNet$^\ast$). The collaboration of shading $S$ and reflectance $R$ priors leads to a substantial decrease in $\triangle E^*$ from 5.12 to 3.55 in the CIELAB color space. Moreover, the design of the mixture of semantic grid experts enables specific experts to focus on individual material categories, resulting in a PSNR of 30.14 dB. The above experiments are conducted with ideal priors. Subsequently, the predictions for $S$, $R$, $M$  from joint RGB-Spectral decomposition model are integrated with HDRNet (denoted as JDM-HDRNet). \cref{tab:jdm_hdrnet}(b) shows that the JDM-HDRNet exhibits comparable performance across three metrics in the $S$ set, indicating that the localized brightness adaptation module demonstrates adaptive adjustment capabilities  in response to shading image degradation. In addition, we observe that the performance of the reflectance ($R$) set remains relatively unaffected, achieving an improvement of 0.98 dB. Because both the Lr-MSI and 16-bit RGB images have undergone consistent transformation into the reflectance space using the same $S$ from the joint decomposition model. Regarding  the material semantic prior (M), JDM-HDRNet dynamically fuses different grid experts at the bilateral grid level to avoid the inaccurate segmentation of $M$. Overall, JDM-HDRNet achieves a gain of 2.08 dB compared to the HDRNet baseline, which demonstrates that the low resolution spectral information can be effectively leveraged by the joint decomposition model.
\begin{table}[t]
\setlength{\abovecaptionskip}{0.0em}
\setlength{\belowcaptionskip}{0.0em}
\caption{Ablation study of the proposed JDM-HDRNet on the $S$, $R$ and $M$ priors. }
\centering
\begin{subtable}[t]{0.47\textwidth}
\caption{Ablation study on the ideal values of $S^\ast$, $R^\ast$ and $M^\ast$ priors. }
\resizebox{!}{0.93cm}{
\setlength{\tabcolsep}{0.5mm}{
\begin{tabular}{c|ccc|ccc}
\hline
\multicolumn{1}{l|}{}                                                             & $S^\ast$ & $R^\ast$ & $M^\ast$ & PSNR↑  & SSIM↑  & $\triangle E^\ast$↓    \\ \hline
HDRNet                                                                            &   &   &   & 27.75 & 0.939 & 5.12 \\ \hline
\multirow{3}{*}{\begin{tabular}[c]{@{}c@{}}JDM-\\ HDRNet$^\ast$ \end{tabular}}                                                     & 
                                                                                \checkmark &   &   & 28.68 & 0.957 & 4.29 \\
                                                        
                                                                                  & \checkmark & \checkmark &   & 29.68 & 0.968 & 3.55 \\
                                                                                  & \checkmark & \checkmark & \checkmark & 30.14 & 0.972 & 3.44 \\ \hline
\end{tabular}
}
}
\end{subtable}
\begin{subtable}[t]{0.47\textwidth}
\centering
\caption{Ablation study on $S$, $R$ and $M$ priors predicted from the joint decomposition model. }
\resizebox{!}{0.93cm}{
\setlength{\tabcolsep}{0.6mm}{
\begin{tabular}{c|ccc|ccc}
\hline
\multicolumn{1}{l|}{}                                                             & $S$ & $R$ & $M$ & PSNR↑  & SSIM↑  & $\triangle E^\ast$↓    \\ \hline
HDRNet                                                                            &   &   &   & 27.75 & 0.939 & 5.12 \\ \hline
\multirow{3}{*}{\begin{tabular}[c]{@{}c@{}}JDM-\\ HDRNet\end{tabular}}                                                       & \checkmark &   &   &  28.59 & 0.953 & 4.27\\
                                                                                  & \checkmark & \checkmark &   &  29.57 & 0.966 & 3.62 \\
                                                                                  & \checkmark & \checkmark & \checkmark &29.83 & 0.967 & 3.60 \\ \hline
\end{tabular}

}
}
\end{subtable}
\label{tab:jdm_hdrnet}
\end{table}
\begin{table}[t!]
\centering
\setlength{\abovecaptionskip}{0.0em}
\setlength{\belowcaptionskip}{0.0em}
\caption{Comparisons of our JDM-HDRNet with previous enhancement methods. }
\fontsize{8}{10}\selectfont
\begin{subtable}[t]{1.0\textwidth}
\setlength{\tabcolsep}{0.7mm}{
\begin{tabular}{c|ccccccccc|c}
\hline
     & \scriptsize{DPE}  & \scriptsize{CSRNet}  & \scriptsize{3D LUT} & \scriptsize{CLUT}  & \scriptsize{SepLUT-S} & \scriptsize{SepLUT-L} & \scriptsize{4D LUT} & \scriptsize{HDRNet}  & \scriptsize{UPE}    & \scriptsize{Ours} \\ \hline
PSNR↑ & 22.81 & 26.34  & 27.52   & 27.30 & 27.57    & 28.08  & 27.77 & 27.75  & 28.19 & \textbf{29.83}      \\
SSIM↑ & 0.806 & 0.923  & 0.926   & 0.938 & 0.933    & 0.944  &  0.935 & 0.939  & 0.946 & \textbf{0.967 }     \\
$\triangle E^\ast$↓   & 11.06 & 6.44   & 5.39   & 4.63  & 5.37    & 4.26  &  4.32   & 5.12   & 4.79  & \textbf{3.60}      \\ \hline
\end{tabular}
}
\end{subtable}
\label{tab:sota}
\end{table}

\subsection{Comparisons with Previous Methods}


\;\;\:\:  \textbf{Quantitative Comparisons.} We compare JDM-HDRNet with previous color transform-based methods for tone enhancement. In \cref{tab:sota}, DPE \cite{chen2018deep} exhibits inferior performance compared to other methods, attributed to its unpaired setting, leading to output images affected by undesired artifacts introduced by the generative adversarial network. Among the color transform functions, the MLPs based CSRNet  \cite{he2020conditional} exhibits a slightly lower PSNR of 26.34 dB. In contrast, the 3D LUTs based methods, such as 3D LUT \cite{zeng2020learning}, CLUT \cite{zhang2022clut}, SepLUT-S \cite{yang2022seplut}, and 4D LUT \cite{liu20234d}, demonstrate the similar PSNR ranging from 27.30 dB to 27.77 dB. Notably, SepLUT-L  \cite{yang2022seplut} with larger parameters exhibits a superior performance because it simultaneously takes advantage of both 1D and 3D LUTs for enhancement, the 1D LUT adjusts image contrast to achieve a more uniform distribution in an image-adaptive manner. Additionally, UPE \cite{wang2019underexposed} estimates an image-to-illumination mapping with constraints of smoothness loss, which is exclusively designed to enhance images under diverse lighting conditions, and it achieves good performance in handling high dynamic range scenes in the Mobile-Spec dataset. Previous methods face challenges in learning accurate tone enhancement for HDR scenes, while our priors-guided JDM-HDRNet outperforms previous methods by effectively leveraging $S$, $R$, $M$ priors predicted from the joint RGB-Spectral decomposition model to provide explicit guidance.


\begin{figure}[t]
  \centering
  \includegraphics[width=0.95\linewidth]{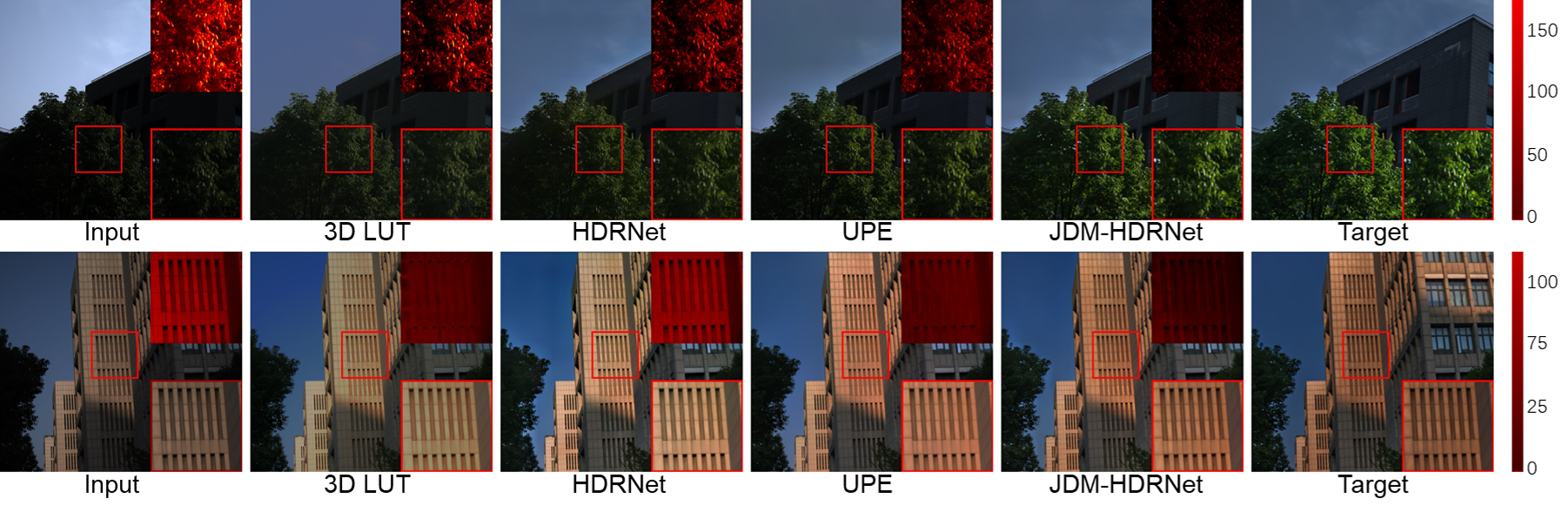}
  \caption{Qualitative comparisons on the Mobile-Spec dataset. The error maps displayed at the top-right corner illustrate the differences with ground truth for each image.
  }
  \label{fig:sota}
\end{figure}


\textbf{Qualitative Comparisons.} \cref{fig:sota} presents qualitative comparisons of two representative examples. In the first row, JDM-HDRNet exhibits enhanced contrast and a wider dynamic range in the plant area, while other methods appear comparatively dim. In the second row, JDM-HDRNet shows reduced color deviation with the ground truth in the wall area, indicating the beneficial role of spectral information integration in improving photographic realism. In summary, JDM-HDRNet, guided by $S$, $R$ and $M$ priors, achieves more accurate and pleasing colors for high dynamic range scenes.



\textbf{Limitations.} The spatial resolution of Lr-MSI in this study is fixed at $16\times 16$, a setting not yet achievable on commercial smartphones. Increasing the spatial resolution would improve the accuracy of the joint decomposition model.  Further exploration is required to address the constrained spectral imaging capabilities in the context of image enhancement and related applications on mobile devices.

\section{Conclusion}
We have investigated the effectiveness of Lr-MSI in the mobile ISP pipeline for enhancing tone enhancement. We first construct the high-quality Mobile-Spec, a pioneering  dataset comprising both aligned spectral and RGB images. Then we analyze the role of Lr-MSI from the perspective of spectral image intrinsic decomposition. Shading, reflectance and material semantic priors are derived using the joint decomposition model to address the inherent complexity of spectral images and the constraints of spectral imaging capabilities. Subsequently, JDM-HDRNet is meticulously designed under the explicit guidance of above priors, which exhibits superior performance than the competitive methods. Our dataset and method are expected to attract further research into this area and explore the untapped potential of spectral information on mobile devices.

\section*{Acknowledgments} 
This research was supported by National Key Research and Development Program of China (2023YFF0713300), National Natural Science Foundation of China under Grant (62071216), Leading Technology of Jiangsu Basic Research Plan under Grant (BK20192003), and National Research Foundation Singapore Competitive Research Program (award number CRP29-2022-0003). We thank the device support from NJU-ZPTECH University-Enterprise Joint Laboratory for Computational Spectral Imaging.

\newpage

\title{Supplementary Material: Joint RGB-Spectral Decomposition Model Guided Image Enhancement in Mobile Photography} 
\author{}
\institute{}
\titlerunning{ }
\authorrunning{ }
\maketitle

\section{More Details about Mobile-Spec Dataset}
\subsection{Mobile-Spec Dataset Overview}
In Figure \ref{fig:dataset}, we present a diverse collection of representative samples from the Mobile-Spec dataset. It can be found that the Mobile-Spec dataset comprises two parts. The first encompasses images captured by the smartphone and hyperspectral camera, while the second comprises images related to shading, reflectance, and material segmentation. The 16-bit input paired with corresponding 8-bit target images are constructed for the task of tone enhancement, predominantly featuring outdoor scenes with high dynamic range. It should be noted that both 16-bit input and  8-bit target images are in the sRGB color space, and we focus on the tone enhancement task rather than learning the whole pipeline from raw data to final output. The hyperspectral images, obtained using the GaiaSky-mini2 \cite{dualixGaiaskyminiHyperspectral}, are intended to overcome the limitations in the spectral imaging capabilities of mobile devices. On the right side of Figure \ref{fig:dataset}, we showcase  shading, reflectance, and material segmentation images for each sample. The shading and material segmentation images serve as targets for the prediction of the joint RGB-Spectral decomposition model. Notably, the shading images are averages of the 850-1000 nm bands in hyperspectral images. It reveals that the near-infrared bands can approximately delineate the distribution of shadows and illumination in outdoor environments. The material segmentation images, on the other hand, are meticulously labeled by human annotators. Our Mobile-Spec dataset aims to establish a high-quality foundation, offering insights and laying the groundwork for further exploration of spectral information in mobile photography.

\subsection{Alignment of Dual-Camera System}
\begin{figure}[H]
  \centering
  \includegraphics[width=1.0\linewidth]{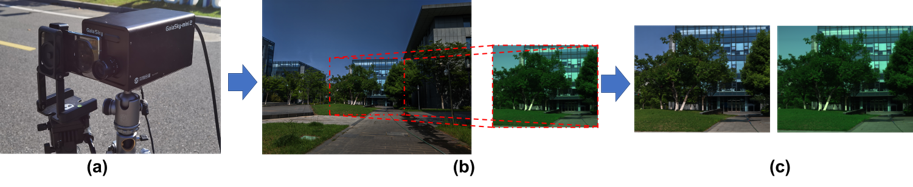}
  \caption{(a) The dual camera system which consists of the high-end commercia smartphone and the GaiaSky-mini2 hyperspectral camera \cite{dualixGaiaskyminiHyperspectral}. (b) Image Matching: the overlapping region is detected by the SIFT descriptor \cite{lowe2004distinctive}, then the affine transformation is performed on the smartphone-captured RGB image to align with the pseudo-RGB image from the hyperspectral data. (c) The aligned pair of RGB and hyperspectral images, the hyperspectral image is transformed into the pseudo-RGB image.
  }
  \label{fig:align1}
  \vspace{-1.5em}
\end{figure}

\;\;\:\:  \textbf{Dual Camera System.} Figure \ref{fig:align1}(a) showcases our dual-camera setup, featuring the high-end commercial smartphone and the GaiaSky-mini2 hyperspectral camera \cite{dualixGaiaskyminiHyperspectral}, which operates on a scanning-based imaging principle to guarantee the quality of the captured hyperspectral images. The two cameras are positioned in close proximity to each other to minimize differences in viewing angles as much as possible. During the capture of each scene, both cameras are set to their default shooting settings and operate synchronously.

\textbf{Image Matching.} As observed in Figure \ref{fig:align1}(b), the RGB images captured by the smartphone own a larger field of view and higher spatial resolution (4096 × 3072 × 3), compared to the hyperspectral images, which feature a smaller field of view and resolution (1057 × 960 × 176). To facilitate the study of tone enhancement tasks, we align the RGB images to the hyperspectral images. Specifically, we first convert the hyperspectral images into the pseudo-RGB format. Then, using the SIFT algorithm \cite{lowe2004distinctive}, we calculate feature points between the smartphone-captured RGB images and the pseudo-RGB images from the hyperspectral data. The feature points are matched to compute the transformation homography matrix between the two images. Finally, using the homography matrix, we perform an affine transformation on the smartphone-captured RGB images, resulting in aligned pairs of RGB and hyperspectral images, as illustrated in Figure \ref{fig:align1}(c).

\begin{figure}[H]
  \centering
  \includegraphics[width=1.0\linewidth]{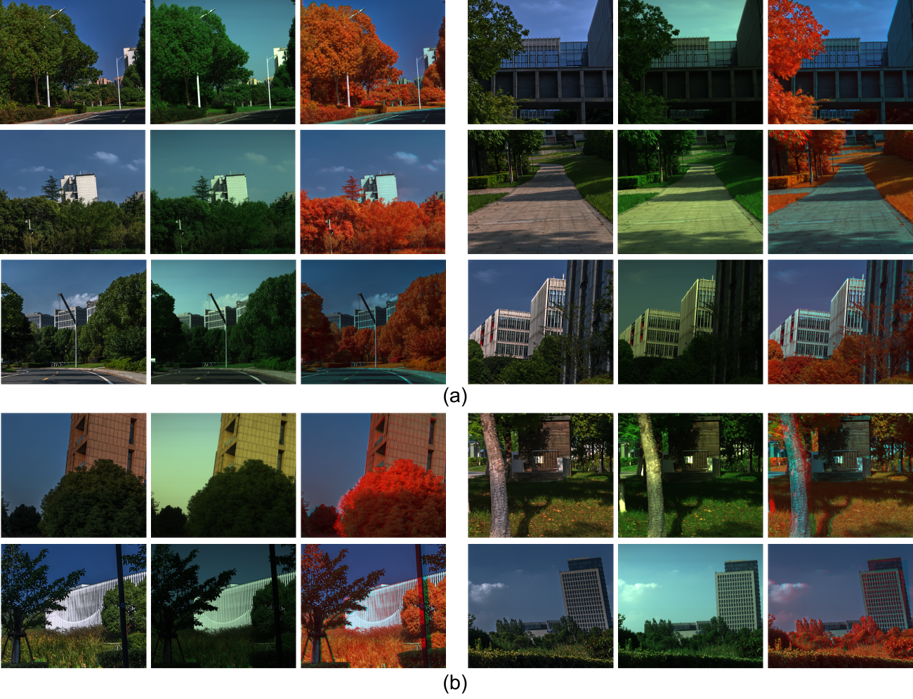}
  \caption{We substitute the R channel in the smartphone-captured RGB image with the shading image, which allows for a visual evaluation of the image registration quality. In each sample, the first column is the 
  smartphone-captured RGB image, the second column is the pseudo-RGB image from the hyperspectral data, the third column is the fused image which replaces the R channel with the shading image. (a) Samples in our Mobile-Spec dataset demonstrate high alignment accuracy. (b) Samples that exhibit artifacts and misalignments are discarded in the dataset filtering procedure.
  }
  \label{fig:align2}
  \vspace{-1.5em}
\end{figure}

\textbf{Filtering of Mismatched Samples.} Due to factors such as depth of field and errors in feature point matching, there can be pixel alignment errors between RGB and hyperspectral image pairs. We filter out samples with significant registration errors. To visually assess the alignment of each sample, we substitute the R channel in the smartphone-captured RGB image with the shading image from the hyperspectral data. This method allows for a visual evaluation of the image registration quality. As shown in Figure \ref{fig:align2}(b), samples that exhibit artifacts and misalignments are excluded. Conversely, samples from our Mobile-Spec dataset demonstrate high alignment accuracy, as displayed in Figure \ref{fig:align2}(a).

Compared to the RGB images (1057 × 960 × 3), the Lr-MSI possesses a markedly limited spatial resolution (16 × 16 × 10). Given that the spatial resolution setting for Lr-MSI is very small, the alignment errors can be negligible in the JDM-HDRNet. We believe that high-precision RGB-hyperspectral image pairs are meaningful for many other tasks, such as joint RGB-spectral pansharpening \cite{loncan2015hyperspectral}, reconstruction \cite{chen2023prior}, segmentation \cite{habili2022hyperspectral}, and illumination estimation \cite{thomas2015illuminant}. Therefore, we ensure that the Mobile-Spec dataset maintains minimal alignment errors.

\subsection{Material Segmentation}
\begin{figure}[H]
  \centering
  \includegraphics[width=1.0\linewidth]{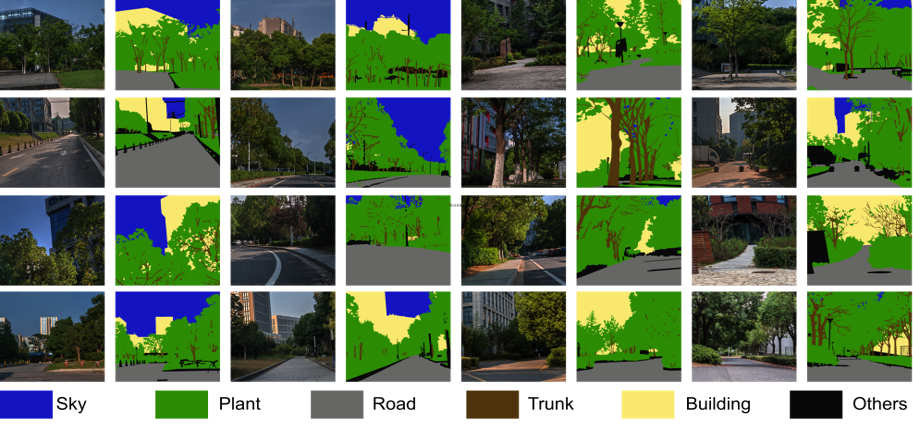}
  \vspace{-2.0em}
  \caption{ The Mobile-Spec dataset is meticulously labeled by human annotators. The segmentation categories are the sky, plant, road, trunk, building, and others, which reflect the most commonly encountered subjects in outdoor scenes.
  }
  \label{fig:seg}
  \vspace{-4.5em}
\end{figure}

Figure \ref{fig:seg} showcases material segmentation images from our dataset, wherein the segmentation categories have been designated as plant, trunk, building, road, sky, and others, reflecting the most commonly encountered subjects in outdoor scenes. It is evident that our semantic segmentation annotations are of exceptionally high granularity, particularly notable in the detailing of individual leaves within the plant area. Our Mobile-Spec dataset is also expected to advance research in the joint analysis of RGB and spectral images for material segmentation.

\subsection{The Approximated Shading Prior }
\begin{figure}[H]
  \centering
  \includegraphics[width=1.0\linewidth]{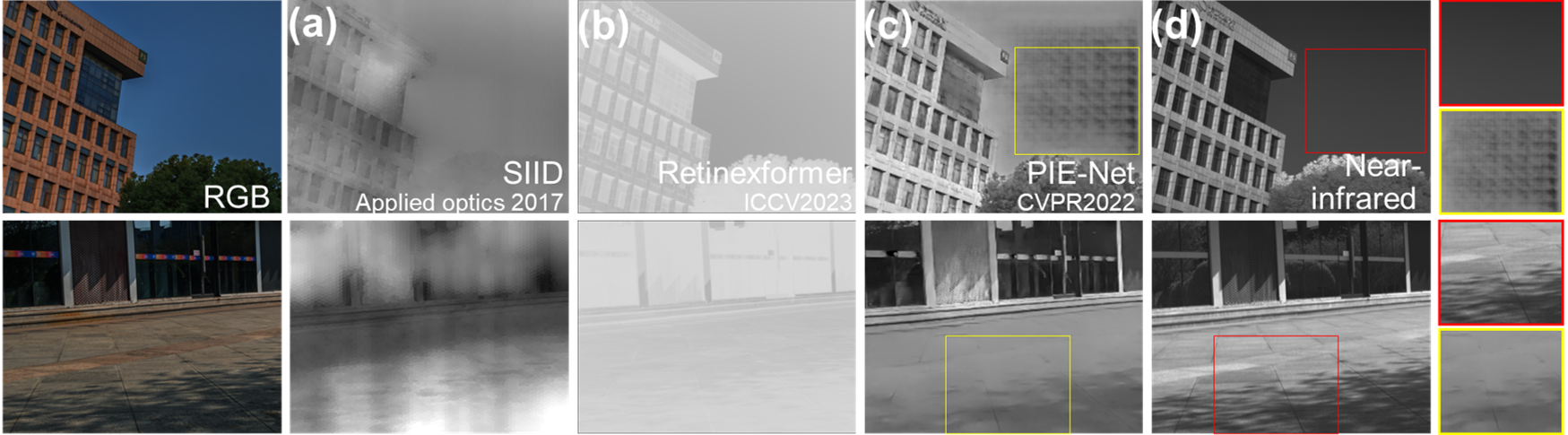}
  \vspace{-1.5em}
  \caption{Since it is difficult to obtain the shading GT for outdoor scenes, we consider four ways to approximate the shading term:  (a) Intrinsic decomposition of hyperspectral images \cite{chen2017intrinsic}. (b) Implicit learning \cite{ cai2023retinexformer}. (c) Intrinsic decomposition of RGB images \cite{das2022pie}.  (d) Near-infrared images \cite{cheng2019non}. The near-infrared images serves as a reliable guide map to  estimate shading.
  }
  \label{fig:shading}
  \vspace{-4.0em}
\end{figure}

Since shading GT of complex outdoor scenes is difficult to obtain, the near-infrared images serve as the guide map to approximate shading instead of real shading GT. We consider four methods for approximating shading: (a) Intrinsic decomposition of hyperspectral images \cite{chen2017intrinsic}: This method fails to handle complex outdoor scenes. (b) Implicit learning (e.g., Retinexformer \cite{cai2023retinexformer}): This method produces results more like the grey image instead of shading. (c) Intrinsic decomposition of RGB images: PIE-Net \cite{das2022pie} relies heavily on training data and lacks interpretability, which is unstable and may produce artifacts and over smoothness. (d) Near-infrared images \cite{cheng2019non}: Figure~\ref{fig:shading} shows near-infrared images stands out as reliable guide map for approximating shading, since they are less sensitive to reflectance variations \cite{cheng2019non}. So we adopt near-infrared images for shading estimation as a simple yet effective way.

\begin{figure}[H]
  \centering
  \includegraphics[width=0.67\linewidth]{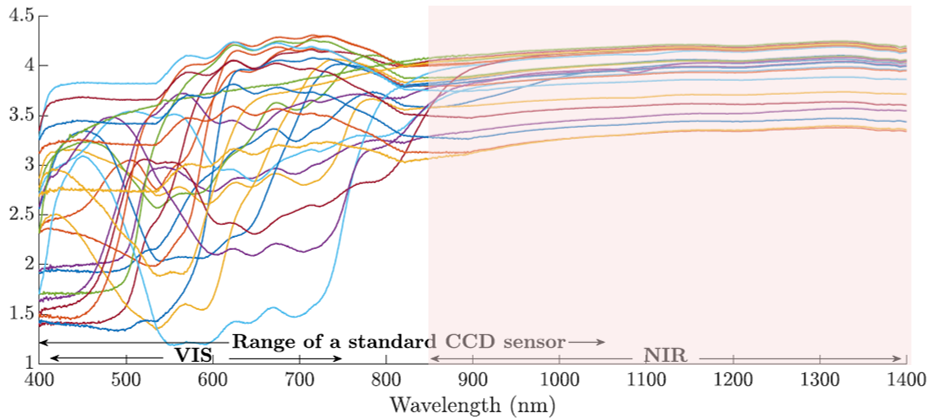}
  \vspace{-1.0em}
  \caption{The reflectance versus wavelength curves for 24 colourants on a colour checker. Different colors exhibit a trend towards flattening and convergence with increasing wavelengths in the near-infrared band. This figure is from \cite{cheng2019non}.
  }
  \label{fig:nir}
  \vspace{-3.0em}
\end{figure}

Figure \ref{fig:nir} delineates the spectral reflectance values sampled across 24 colorants on a color checker. The spectral curves exhibit a trend towards flattening and convergence with increasing wavelengths, suggesting the near-infrared spectrum's viability as a proxy for shading \cite{cheng2019non}. It is pertinent to note that the precise estimation of shading is not the focal point of this paper; rather, our objective is to harness the shading prior approximated by near-infrared bands to enhance the image quality in tone enhancement tasks.

\begin{figure}[H]
  \centering
  \includegraphics[width=0.9\linewidth]{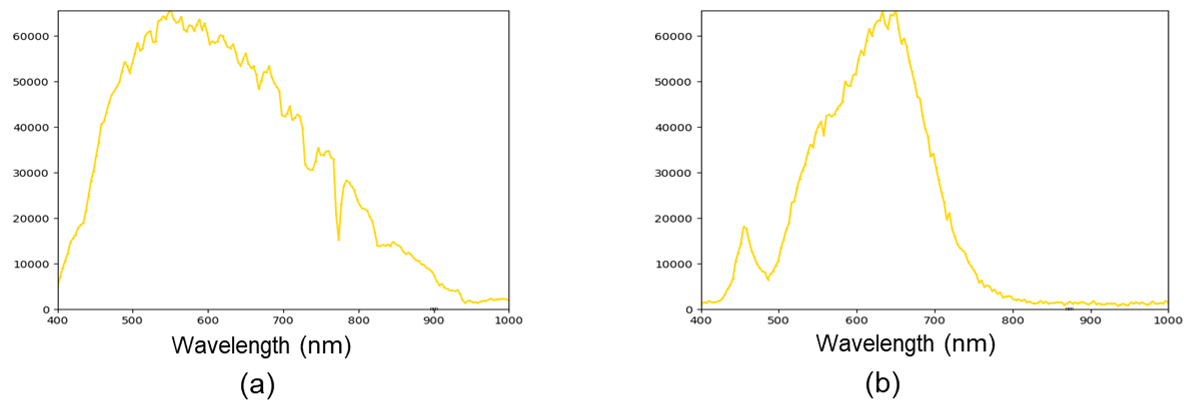}
  \vspace{-1.0em}
  \caption{(a) Sunlight. (b) LED light source. The spectral curves of sunlight and LED light source are captured by our hyperspectral camera with the white board. The spectrum intensity of LED light source  swiftly diminishing at near-infrared wavelengths. While the sunlight  exhibits a broad and continuous spectrum. 
  }
  \label{fig:colorboard_curve}
\end{figure}

 As illustrated in Figure~\ref{fig:colorboard_curve}, the spectral curves of sunlight and LED light source are captured by our hyperspectral camera with the white board. The sunlight encompasses a broad and continuous spectrum across the visible range, extending into the ultraviolet (UV) and infrared (IR) regions, with its intensity being relatively uniform across the visible spectrum, albeit with minor variations. Conversely, the spectrum of the LED light source is characterized by a pronounced peak in the blue region, with its intensity swiftly diminishing at near-infrared wavelengths.

\begin{figure}[H]
  \centering
  \includegraphics[width=1.0\linewidth]{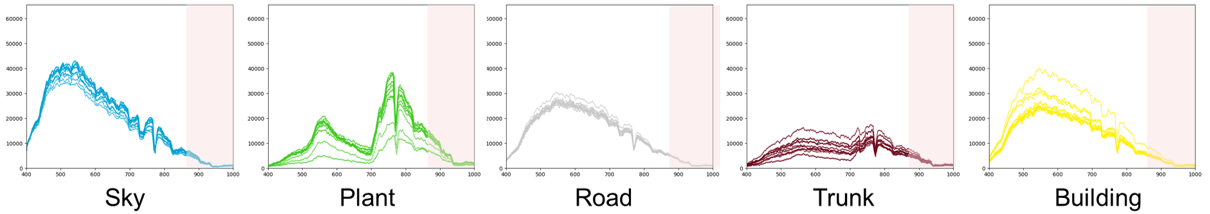}
  \vspace{-2.0em}
  \caption{Under the single outdoor illumination of sunlight, the spectral curve trends of different materials in the near-infrared spectrum (850-1000nm)  tend to be consistent. 
  }
  \label{fig:material_curve}
  \vspace{-6.0em}
\end{figure}

\begin{figure}[H]
  \centering
  \includegraphics[width=0.9\linewidth]{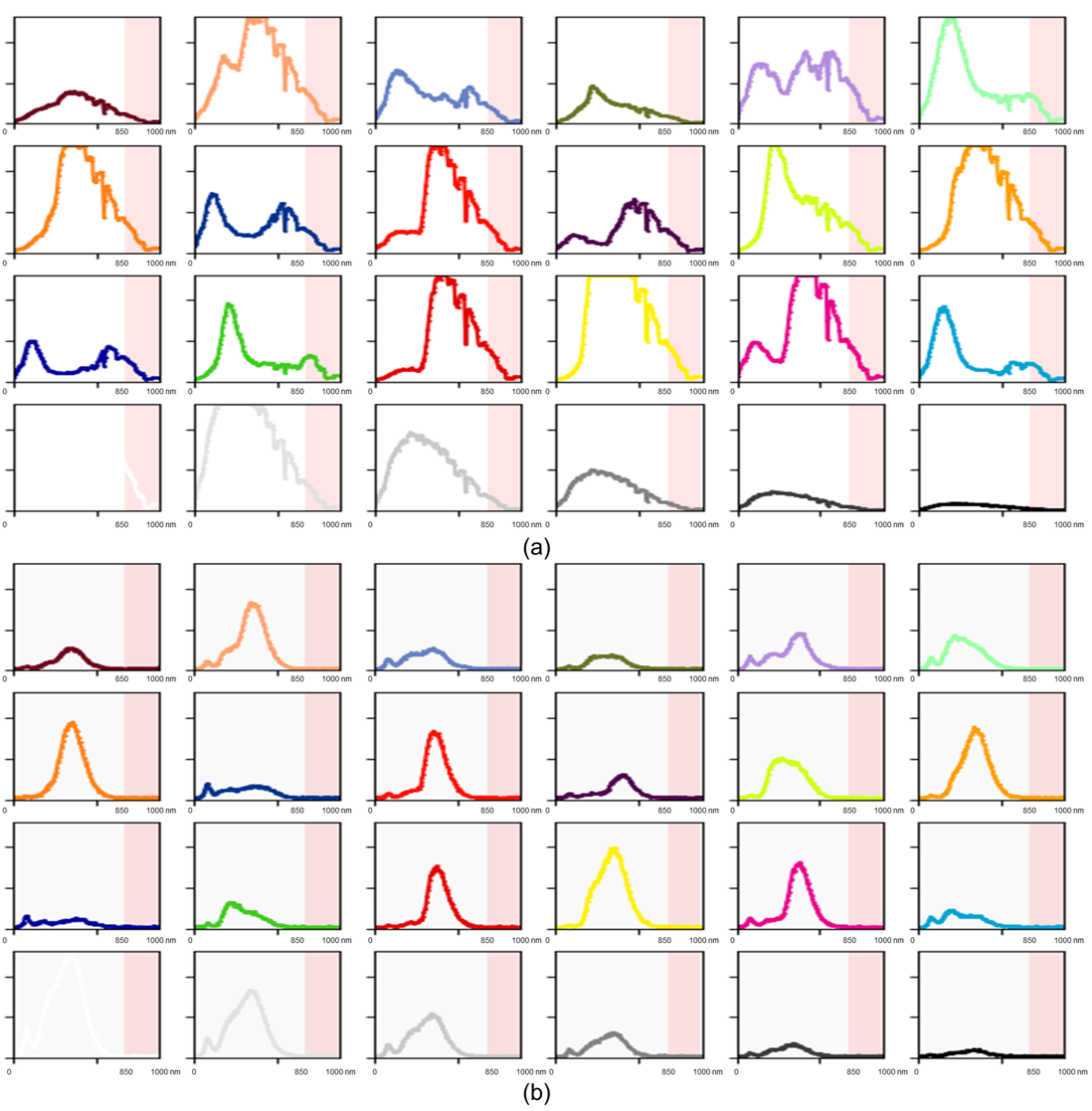}
  \vspace{-1.0em}
  \caption{(a) Sunlight. (b) LED light source. The spectral curves of 24 distinct colors on a standard color checker under illuminations of sunlight and LED light source are captured by our hyperspectral camera. 
  }
  \label{fig:colorboard_curve2}
\end{figure}

Furthermore, we have drawn spectral curves of 24 distinct colors on a color checker using our hyperspectral camera under two light sources. As evident from Figure~\ref{fig:colorboard_curve2}(a),  for the illumination of sunlight within the visible light spectrum (400-750nm), different colors display unique spectral curve characteristics. However, in the near-infrared spectrum (850-1000nm), the spectral curves of different colors tend to converge, aligning with the conclusion drawn in Figure \ref{fig:nir}. However, under LED illumination, various colors demonstrate no significant response within the near-infrared spectrum. Given that most indoor lighting sources are LED lights, the assumption of using the near-infrared spectrum as a proxy for shading does not hold indoors. Consequently, the samples in our Mobile-Spec dataset are exclusively captured in outdoor settings. As shown in Figure~\ref{fig:material_curve}, under the single outdoor illumination of sunlight, the spectral curve variations of different materials in the near-infrared spectrum (850-1000nm) tend to be consistent. Notably, the divergence in spectral response, both among various materials and across distinct pixels of the same material, is more pronounced in terms of intensity variation. This suggests  the near-infrared band may serve as an approximation for the shading prior in the outdoor environment.

\begin{table*}[h]
\centering
\caption{ Objective quality assessment of tone-mapped images  across four representative datasets. Our Mobile-Spec dataset achieves comparable structural fidelity with PPR10K \cite{liang2021ppr10k} and MIT FiveK \cite{bychkovsky2011learning}.  } 
\setlength{\tabcolsep}{1.5mm}{
\begin{tabular}{c|ccc|c}
\hline
                    & HDR+ \cite{hasinoff2016burst} & FiveK \cite{bychkovsky2011learning} & PPR10K \cite{liang2021ppr10k} & Mobile-Spec \\ \hline
Structural Fidelity↑ \cite{yeganeh2012objective} & 0.130 & 0.150 & \bf{ 0.193}   & 0.176      \\ \hline
\end{tabular}
}


\label{sf}
\end{table*}

\begin{table}[h]
\centering
\caption{Comparisons with previous hyperspectral datasets.  } 
\setlength{\tabcolsep}{1.5mm}{
\begin{tabular}{c|cccc|c}
\hline
                    & Harvard \cite{chakrabarti2011statistics}    & CAVE \cite{choi2017high}  & KAIST \cite{yasuma2010generalized}        & PaviaU    & Mobile-Spec     \\ \hline
scenes              & 75      & 32        & 30                 & 1         & 200      \\
spatial (x-y)   & 1392×1040 & 512×512 & 2704×3376  & 610×340 & 1057×960 \\
spectral ({\scriptsize$\lambda$,nm})  & 420-720 & 400-700   & 400-700     & 430-860   & 400-1000 \\
channels            & 30      & 31        & 31              & 103       & 176      \\
segmentation        & ×       & ×         & ×           & \checkmark     & \checkmark   \\
aligned RGB           & ×       & ×         & ×                   & ×         & \checkmark       
      \\ \hline
\end{tabular}
}

\label{tab3}
\end{table}

\subsection{8-bit Target Image}
Since tone-mapped targets are highly subjective, as different individuals have varying aesthetic preferences, we employ a commercial privacy model to generate targets of Mobile-Spec dataset that integrate the aesthetics of multiple experts.  This model is trained by a large-scale and high-quality commercial dataset, which is elaborately adjusted by professional photographers and artists. Then we meticulously filter the target images through subjective assessments, considering factors such as chromatic aberration, sharpness, noise and artifacts. Samples that do not meet the criteria are excluded.
We adopt the structural fidelity term in TMQI \cite{yeganeh2012objective} to evaluate the objective quality of the Mobile-Spec dataset. It should be noted that we only reserve the structural fidelity term and the statistical naturalness term is removed, since the tone enhancement images in  Mobile-Spec are high dynamic range scenes, which do not comply with statistical naturalness of common pictures. Table~\ref{sf} shows the Mobile-Spec dataset achieves comparable structural fidelity with PPR10K \cite{liang2021ppr10k} and MIT FiveK \cite{bychkovsky2011learning}, highlighting its potential as a solid foundation for exploring the role of Lr-MSI in the tone enhancement task.

\subsection{Comparisons with Previous Hyperspectral Datasets}

From Table~\ref{tab3}, it can be seen that our Mobile-Spec dataset encompasses more diverse scenes and a broader spectral range compared to previous hyperspectral datasets. Additionally, the Mobile-Spec dataset owns high resolutions in both spatial and spectral dimensions. We provide aligned RGB images captured by smartphones and meticulously labeled segmentation maps. Consequently, the Mobile-Spec dataset is anticipated to contribute to related fields such as hyperspectral reconstruction, segmentation, and pansharpening.

\section{Qualitative Comparisons}
\subsection{Test Set of Mobile-Spec}
\begin{figure}[H]
  \centering
  \includegraphics[width=1.0\linewidth]{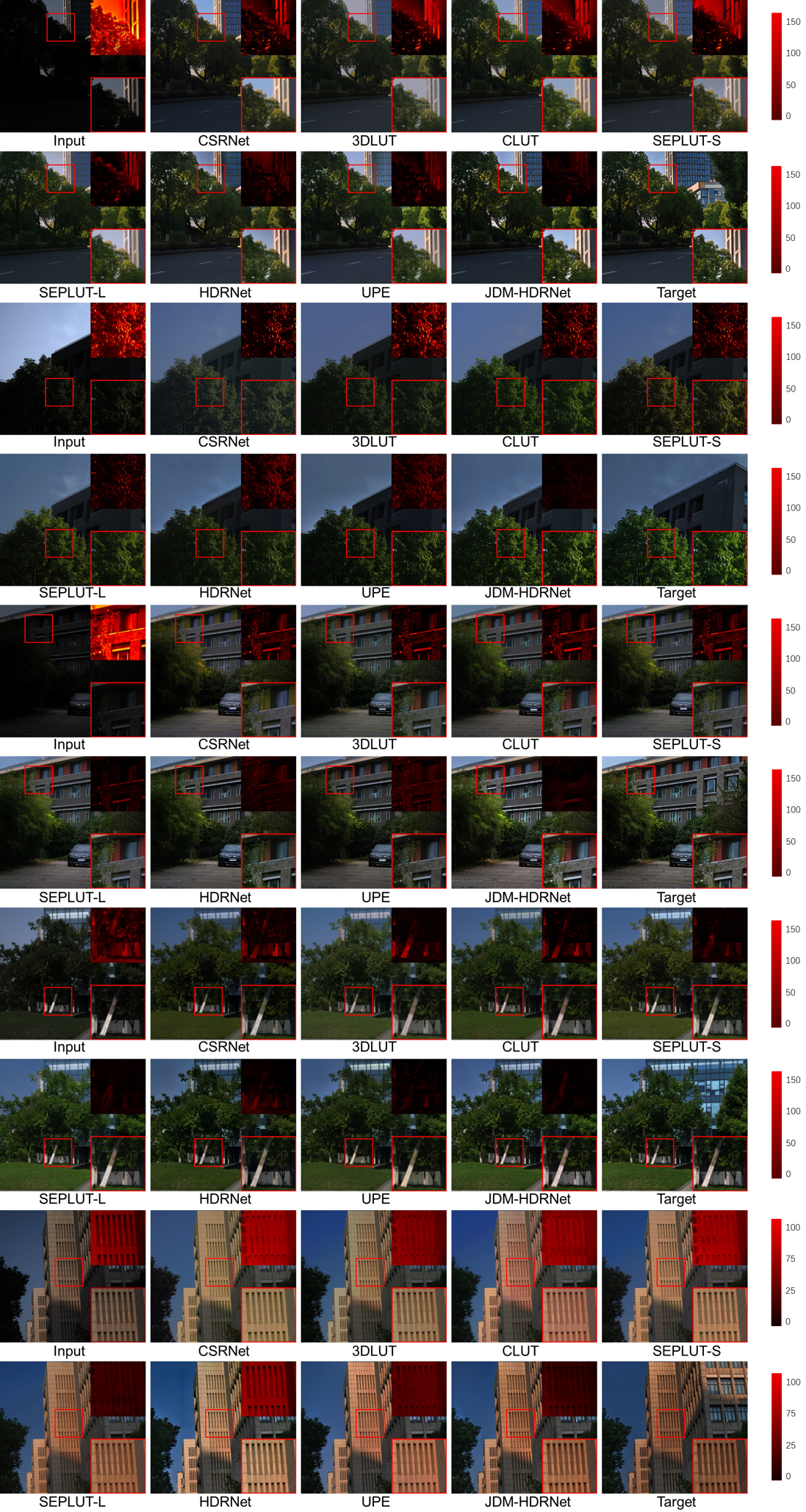}
  \vspace{-2.0em}
  \caption{More qualitative comparisons on the Mobile-Spec dataset.   }
  \label{fig:vis_sp}
\end{figure}

Figure~\ref{fig:vis_sp} presents additional  comparative visualization of the results between JDM-HDRNet and other methods. From these samples, we can draw the following conclusions: (1) In areas with drastic local brightness variations, such as sunlight-dappled foliage exhibiting diverse light and shadow distributions, JDM-HDRNet adaptively accommodates high-dynamic-range scene tone adjustments by isolating the shading component. (2) Incorporating the reflectance prior of Lr-MSI with an expanded color channel, and designing a specialized grid expert for each material enable JDM-HDRNet to produce tone-enhanced outcomes with minimal color deviation. For instance, the exterior walls of the building in the fifth sample and the red wall in the third sample. Given that the Mobile-Spec dataset predominantly features sky and plant, cooler tones like blue and green dominate most colors, leading to inadequate learning for mapping warmer tones such as red. Introducing additional Lr-MSI can mitigate the imbalance in learning across different colors to some extent. JDM-HDRNet achieves more accurate and aesthetically pleasing colors compared to the competitive methods. The qualitative comparison underscores the efficacy of decomposing Lr-MSI into three components: shading, reflectance, and material semantic priors. This framework offers explicit guidance for tone enhancement and overcomes the intrinsic complexity of spectral images. Through the exploration of Lr-MSI in the tone enhancement task, we aim to lay the foundation for the broader application of spectral information in mobile photography.

\subsection{Unseen Samples}
\begin{figure}[H]
  \centering
  \includegraphics[width=0.7\linewidth]{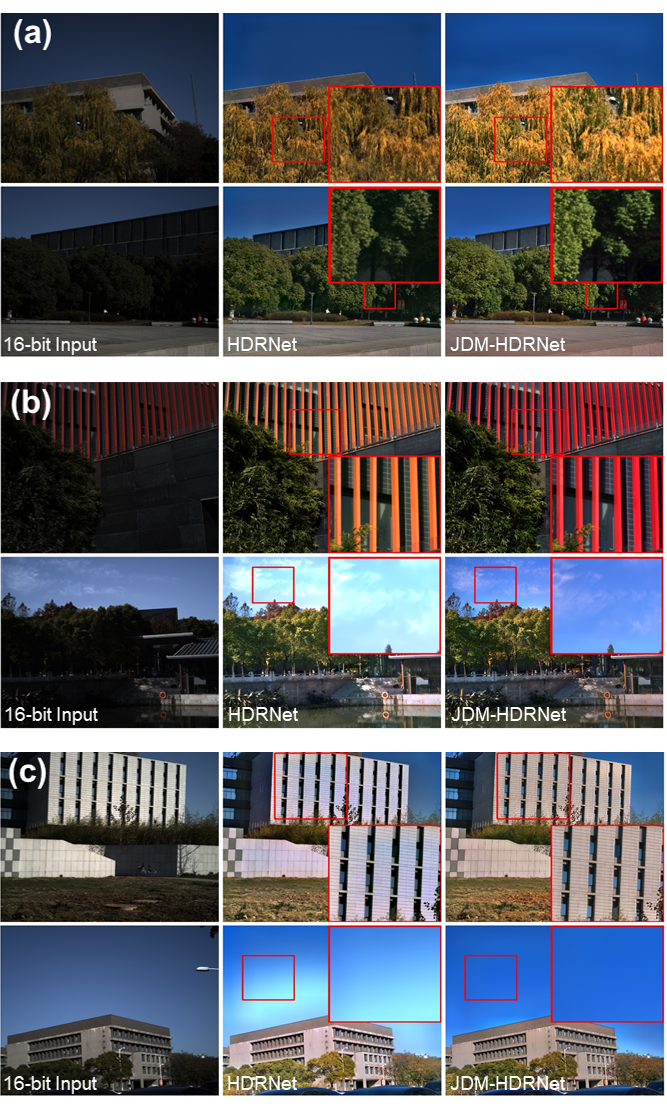}
  \caption{ Evaluations on unseen samples captured from new locations, which contain scenes out of the Mobile-Spec dataset (e.g., pool, yellow leaves). Benefits of Lr-MSI: (a) $S$: enhanced dynamic range; (b) $R$: more accurate color; (c) $M$: context consistency. Explicit priors help constrain JDM-HDRNet outputs, enhancing generalization and robustness on unseen images.
  }
  \label{fig:vis_unseen}
\end{figure}

 To validate the generalization, we capture extra unseen test samples from entirely new locations, which contain scenes outside  of Mobile-Spec (e.g., pool, yellow leaves). Qualitative results in Figure~\ref{fig:vis_unseen} show JDM-HDRNet generates more vivid images with a natural appearance  than HDRNet. The benefits of Lr-MSI lie in the following aspects: (a) Enhanced dynamic range with {$S$}: Shading prior enhances adaptability in dealing with localized brightness variations. (b) More accurate color with {$R$}: JDM-HDRNet generates more realistic color than HDRNet on unseen images.  (c) Context consistency with {$M$}: Introducing {$M$} prior reduces color inconsistency within the same context region. Explicit priors help constrain JDM-HDRNet outputs, enhancing generalization and robustness on unseen images.
 
 \newpage
 
\begin{figure}[H]
  \centering
  \includegraphics[width=1.0\linewidth]{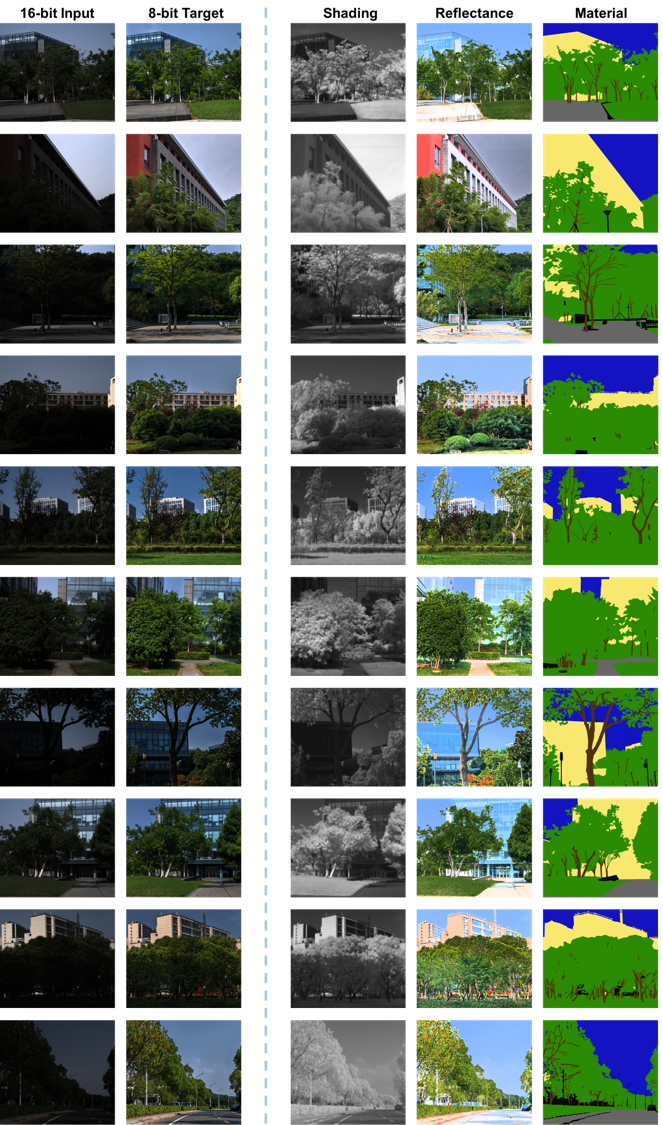}
  \caption{More visualized samples of the Mobile-Spec dataset. The 16-bit RGB images are linear tone-mapped for visualization. 
  }
  \label{fig:dataset}
\end{figure}

\newpage

%
%

\newpage

\bibliographystyle{splncs04}
\bibliography{main}
\end{document}